# Uncertainty-Aware Extreme Point Tracing for Weakly Supervised Ultrasound Image Segmentation

Lei Shi[1,2], Gang Li[3], Junxing Zhang[1,*]

[1] College of Computer Science, Inner Mongolia University, Hohhot, China.

[2] Baotou Medical College, Baotou, China.

[3] Department of Ultrasound Medicine, The Second Affiliated Hospital of Baotou Medical College, Baotou, China.

Email: shilei@mail.imu.edu.cn; junxing@imu.edu.cn

**Abstract:** Automatic medical image segmentation is a fundamental step in computer-aided diagnosis, yet fully supervised approaches demand extensive pixel-level annotations that are costly and time-consuming. To alleviate this burden, we propose a weakly supervised segmentation framework that leverages only four extreme points as annotation. Specifically, bounding boxes derived from the extreme points are used as prompts for the Segment Anything Model 2 (SAM2) to generate reliable initial pseudo labels. These pseudo labels are progressively refined by an enhanced Feature-Guided Extreme Point Masking (FGEPM) algorithm, which incorporates Monte Carlo dropout-based uncertainty estimation to construct a unified gradient–uncertainty cost map for boundary tracing. Furthermore, a dual-branch Uncertainty-aware Scale Consistency (USC) loss and a box alignment loss are introduced to ensure spatial consistency and precise boundary alignment during training. Extensive experiments on two public ultrasound datasets, BUSI and UNS, demonstrate that our method achieves performance comparable to, and even surpassing fully supervised counterparts while significantly reducing annotation cost. These results validate the effectiveness and practicality of the proposed weakly supervised framework for ultrasound image segmentation.

1. Introduction

　　Medical image segmentation is a fundamental task of computer-aided diagnosis and treatment planning[1][2]. However, obtaining high-quality pixel-wise annotations remains expensive and time-consuming, as it requires the expertise of experienced clinicians. For modalities such as ultrasound, MRI, or CT, delineating a single lesion may take several minutes to tens of minutes, making large-scale annotation impractical. Moreover, pixel-level labeling is not only labor-intensive but also prone to inter-observer variability, further limiting its reliability in clinical workflows.

In contrast, weakly supervised annotations, such as bounding boxes or extreme points, can be collected within seconds and thus provide a more feasible alternative in practice. These coarse labels align more naturally with routine clinical annotation habits, where physicians often prefer marking sparse cues rather than carefully tracing full contours. At the same time, the demand for large annotated datasets to train deep segmentation models, such as U-Net based networks[3][4][5], poses a scalability bottleneck under limited resources. Weakly supervised strategies effectively address this conflict by significantly reducing annotation costs while enabling the construction of larger training sets, thereby improving model generalization and bringing automated segmentation closer to clinical deployment.

Nevertheless, existing weakly supervised segmentation methods still suffer from three critical challenges. First, the quality of pseudo labels is often unstable. Approaches based on bounding boxes or extreme points typically rely on heuristic propagation strategies such as random walker [6] or CRF [7], which tend to produce pseudo masks with blurry boundaries, missing details, and considerable noise. Such noisy supervision can accumulate during training and mislead model optimization. Second, most existing works neglect uncertainty modeling when exploiting weak annotations [8-10]. As a result, predictions in ambiguous boundary regions or low-contrast tissues remain inaccurate, and the distinction between high-confidence and uncertain regions is rarely leveraged to guide effective learning. Third, current approaches often make isolated use of point cues or bounding-box cues [7,11], without integrating them in a complementary manner. This underutilization of available supervision limits the geometric constraints provided by extreme points as well as the regional guidance from bounding boxes.

To overcome these limitations, there is an imperative need for a weakly supervised segmentation framework that not only derives reliable pseudo labels from minimal annotations but also incorporates uncertainty estimation to handle ambiguous regions and leverages complementary weak cues in a unified manner. In this work, we present a weakly supervised segmentation framework that requires only four extreme points as annotation for each image, making it highly annotation-efficient compared with pixel-level labeling. From these four points, bounding boxes are automatically constructed and used as prompts for the Segment Anything Model 2 (SAM 2), through which reliable initial pseudo labels are generated without any dense mask supervision. This design combines the advantages of foundation model priors with minimal annotation effort, providing a practical and scalable solution for medical image segmentation tasks.

To further refine the pseudo labels during training, we introduce an Uncertainty-Aware Feature-Guided Extreme Point Masking (UA-FGEPM) algorithm. Specifically, Monte Carlo dropout is applied at inference to obtain multiple stochastic feature predictions, from which both mean response and pixel-wise variance are calculated. The mean feature map is processed by the Sobel operator to highlight boundary gradients, while the variance forms an uncertainty map. These two cues are integrated into a unified cost matrix that guides Dijkstra-based path tracing between extreme points, enabling pseudo labels to be iteratively updated at fixed training intervals. In this way, pseudo labels are progressively optimized to provide more reliable supervision, thereby improving segmentation performance.

Moreover, to enhance training stability, we adopt a dual-branch multi-scale prediction strategy with an Uncertainty-aware Scale Consistency (USC) loss. Unlike conventional consistency regularization, USC selectively emphasizes predictions in high-confidence regions while down-weighting ambiguous or noisy areas, thus preventing error propagation. In parallel, we introduce a box alignment loss that constrains the predicted mask within the annotated bounding box, thereby reinforcing spatial consistency and preventing boundary drift beyond the weak supervision signals.

Overall, our framework unifies foundation model knowledge, uncertainty-aware pseudo-label refinement, and spatial consistency regularization into a coherent weakly supervised design. Specifically, SAM2 with extreme-point bounding boxes provides high-quality initial cues, UA-FGEPM progressively refines pseudo labels through uncertainty-guided iterations, and the combination of USC loss and box alignment loss further enforces spatial consistency during training. This results in an annotation-efficient and model-agnostic solution that achieves competitive performance with fully supervised methods while requiring only four extreme points per image. Fig.1 shows the motivation and segmentation results of our method.

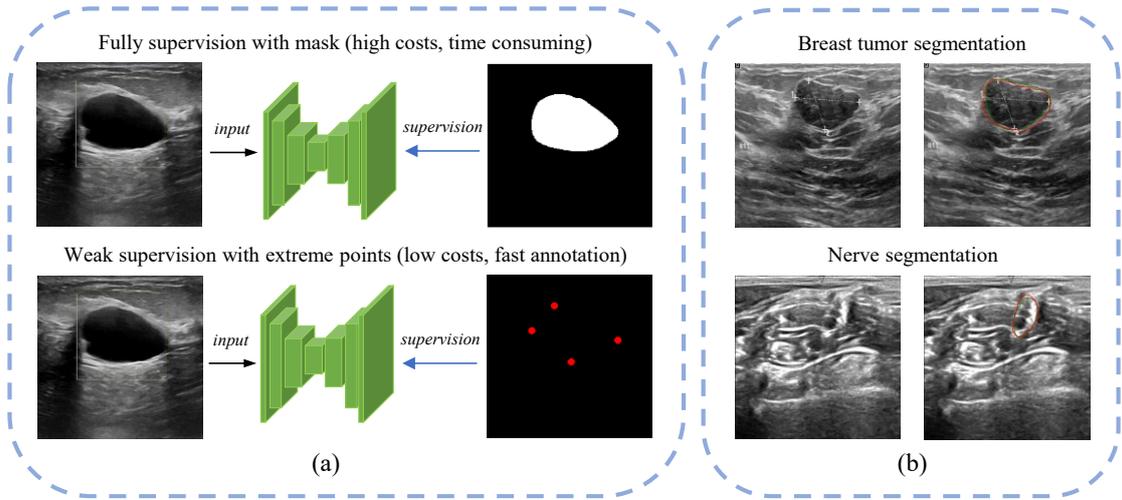

Fig. 1. (a) Fully supervision method versus extreme points supervision method; (b) Two visualization samples with our method. Left column: original images, right column: our segmentation results. The ground truths are in green, and our results are in red. Best zoomed in and viewed in color.

The main contributions of this work are summarized as follows:

- We introduce a simple yet effective strategy that transforms four extreme points into bounding-box prompts for SAM2, enabling the generation of high-quality pseudo labels without any pixel-level annotations. This design substantially reduces annotation effort while exploiting the representational knowledge of foundation models.

- To overcome the limitations of static pseudo labels, we propose an iterative refinement mechanism guided by gradient and uncertainty cues estimated via Monte Carlo dropout. This process progressively corrects noisy or coarse boundaries and enhances the reliability of supervision throughout training.

- We develop a dual-branch prediction architecture that incorporates an

uncertainty-aware scale consistency loss to emphasize reliable regions and suppress uncertain ones, ensuring stable optimization across scales. In addition, a box alignment loss enforces geometric consistency between the predicted masks and the annotated bounding boxes, serving as a structural constraint that improves boundary precision under weak supervision.

This paper is a substantial extension of our preliminary work accepted in GLOBECOM 2025 e-health track [30], where we proposed FGEPM algorithm and iterative training. In this extension, we further use box prompt and SAM2 to generate pseudo labels and introduce uncertainty to improve FGEPM algorithm and design a new scale consistency loss function. We further validate our method on the newly introduced DDTI dataset to demonstrate its strong generalization capability. Compared with state-of-the-art weakly supervised methods, our approach achieves comparable or even superior performance to fully supervised counterparts across all three datasets.

2. Related Work

2.1 Weakly Geometry-Supervised Image Segmentation

Weakly supervised image segmentation aims to segment images with limited annotations, such as bounding boxes, extreme points, and scribbles. Among them, methods using the bounding boxes account for a large proportion. J. Wei et al. [11] propose the mask-to-box (M2B) transformation and a scale consistency (SC) loss to improve polyp segmentation performance. J. Wang and B. Xia [12] integrate a multiple instance learning strategy based on polar transformation to assist image segmentation. Methods using scribbles are also proposed. For example, Q. Chen and Y. Hong [13] propose a scribble-based volumetric image segmentation architecture with a label propagation module. However, bounding boxes enclose entire rectangular areas, inevitably introducing background noise to labels. Scribbles are hand-drawn rough lines that lack precise boundary information. Instead, we choose extreme points as the annotation for weakly supervised learning in this paper.

2.2 Extreme points for image segmentation

Extreme points for segmentation can generally be divided into two categories. In the first category, extreme points are adopted as additional information to guide the segmentation. K.K. Maninis et al. [14] use extreme points as an additional channel input to the CNN and annotate an extra point in the erroneous area for inaccurate cases to recover performance. Z. Wang et al. [15] make interactive annotations by incorporating user clicks on the extreme boundary points. However, the above methods have high model complexity and transferring these methods to medical tasks is challenging. In the second category, extreme points are typically used to supervise the network for training. H. Lee et al. [8] propose a point retrieval algorithm using extreme points for natural image instance segmentation. The method in [6] uses pseudo-mask generation with a random walker algorithm and minimal user interaction for medical image segmentation. R. Dorent et al. [7] use deep geodesics to connect extreme points and employ a CRF regularized loss for 3D Vestibular Schwannoma segmentation.

2.3 Consistency Regularization and Uncertainty-Aware Segmentation

Consistency regularization remains a central paradigm for semi/weakly supervised segmentation, where predictions are encouraged to be invariant across perturbations or scales. Recent advances explicitly address scale variation and representation drift: for example, DEC-Seg [16] enforces dual-scale and cross-generative consistency with dedicated cross-level aggregation/fusion modules and reports gains on multiple medical benchmarks, highlighting the utility of cross-scale agreement in medical images. A parallel trend incorporates uncertainty into consistency to mitigate error propagation in ambiguous regions. Evidential Inference Learning [17] integrates Dempster–Shafer evidence to obtain single-pass uncertainty estimates while regularizing with consistency, and has shown strong performance across medical datasets. Beyond evidential formulations, uncertainty-aware consistency has been pursued via entropy/variance-guided weighting and dynamic curricula; representative examples include UAC [18] and DyCON [19], which modulate consistency by voxel/pixel uncertainty and couple it with contrastive objectives. Entropy-guided contrastive learning [20] leverages pixel-wise entropy to weight contrastive objectives, thereby enhancing feature discrimination while suppressing unreliable supervision in semi-supervised medical image segmentation. Taken together, these findings support uncertainty-aware cross-scale consistency designs that enforce agreement on reliable regions while down-weighting high-entropy/variance pixels, reducing the risk of amplifying label noise.

2.4 Pseudo-Label Refinement with Foundation Models and Dropout-Based Uncertainty

Promptable foundation models are increasingly used to bootstrap pseudo-labels from economical prompts (points/boxes) and then refined during training. SAM2 [21] improves prompt-driven segmentation with a streamlined transformer and streaming memory. MedSAM [22] demonstrates broad medical transfer for promptable segmentation.

After obtaining initial pseudo-labels from foundation models, many works carry out refinement to correct errors, reduce prompt sensitivity, or fuse complementary predictions. CPC-SAM [23] imposes cross-prompt consistency by employing dual branches with different prompts and regularizing prompt invariance. Others adopt iterative correction / prompt updating, such as Iterative Correction Learning [24], which repeatedly refines pseudo labels by re-prompting and entropy regularization. In domain adaptation or semi-supervised settings, fusion strategies combine traditional pseudo-labels with SAM outputs, as seen in SRPL-SFDA[25] and RESAMPL-UDA [26], where consensus among multiple SAM outputs or perturbation-based agreement is used to screen refined pseudo labels. Classic graph/geometry post-processing (connectivity, curvature smoothing) remains effective, exemplified by connectivity-driven pseudo-labeling [27], which helps stabilize mask continuity and suppress spurious artifacts.

Several recent works adopt Monte Carlo Dropout (MC-Dropout) or analogous stochastic forward passes to estimate pixel-wise uncertainty and filter or weight pseudo-labels accordingly. Zeevi et al. [28] propose a frequency-domain extension of MC-Dropout to better capture structural noise and improve uncertainty estimation in segmentation tasks. Wu et al. [29] design UPL-SFDA, which duplicates prediction heads to simulate multiple stochastic predictions and uses inter-head variance to identify reliable pseudo-labels in a source-free adaptation setting.

## 3. Proposed Method

### 3.1 Overall Framework

We propose a weakly supervised segmentation framework that leverages extreme-point annotations as the only supervision. In the initial stage, SAM2 is guided by bounding-box prompts derived from four extreme points to produce pseudo labels of higher quality than coarse labels. During training, the predicted masks are further constrained by enforcing alignment with the ground-truth tight bounding boxes, thereby maintaining structural consistency. To enhance robustness, a dual-branch multi-scale prediction strategy is adopted, where an entropy-based uncertainty-aware scale consistency loss selectively applies stronger regularization to confident regions while down-weighting uncertain areas. In addition, pseudo labels are iteratively refined through an improved UA-FGEPM algorithm, which incorporates Monte Carlo dropout to build a cost matrix that combines both gradient and uncertainty information. This design allows the contours to be traced more precisely from the extreme points and ensures that the pseudo labels become progressively more accurate as training proceeds. Notably, the proposed framework is model-agnostic and can be seamlessly integrated into any segmentation backbone for medical imaging tasks. Fig.2 shows the framework of the proposed Uncertainty-Aware Extreme Point Tracing (UA-EPT) method.

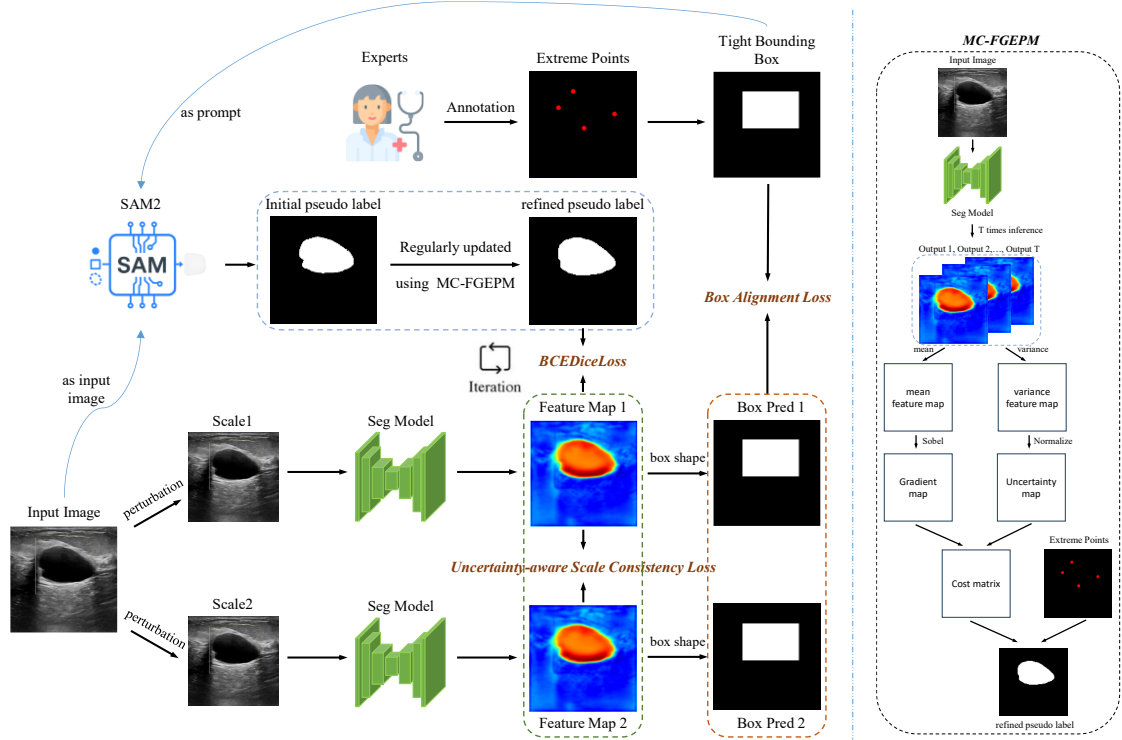

Fig.2 The framework of our UA-EPT.

3.2 Initial pseudo-label generation

In this study, lesion regions in the datasets are intended to be annotated by professional physicians using extreme points, specifically the topmost, bottommost, leftmost, and rightmost points. Throughout the training process, these extreme point annotations serve as the sole supervisory signal. Nevertheless, since the adopted datasets are publicly available and generally provide pixel-level masks rather than extreme point annotations, the required extreme points are systematically derived from the ground-truth masks in a manner consistent with the annotator's perspective. It should be underscored that the ground-truth masks are not employed in any subsequent training stage, thereby ensuring strict adherence to the weakly supervised learning paradigm.

However, the four extreme points alone are insufficient to serve as pseudo labels for guiding the model during the initial training stage. While these points can be used to construct coarse pseudo labels—such as a polygon formed by connecting the points in a counterclockwise order or a tight bounding box derived from them — such representations inevitably contain noise and deviate significantly from the ground truth. Directly adopting them as initial pseudo labels may hinder the improvement of weakly supervised segmentation performance. In our previous work [30], we attempted to generate pseudo labels by extracting lesion contours directly from the original images. Nevertheless, this approach did not fully exploit the powerful 2D segmentation capability of the Segment Anything Model (SAM) and the rich prior knowledge embedded in its pre-trained weights.

SAM2 [21] is explicitly designed for "segment anything" and demonstrates strong generalization with diverse prompt types such as points and boxes. In this work, we utilize SAM2 to directly generate initial pseudo labels under weak prompts. The weak prompts are tight bounding boxes derived from four extreme points, which incur no additional annotation time or cost. Although SAM2 can segment lesions without any prompts, using bounding box prompts improves the segmentation accuracy and brings the results closer to the ground truth. As shown in Fig.2, we leverage the pretrained weights of SAM2 to perform inference on the training images, while using bounding box prompts to constrain the model to segment only within the bounding box region. The segmentation results are employed as initial pseudo labels to guide the training process prior to the first round of pseudo label updates.

3.3 Uncertainty-aware Scale Consistency Loss & Box Alignment Loss

Consistency regularization has been widely adopted in semi-supervised and weakly supervised segmentation [11,16] to improve generalization by enforcing prediction stability under different perturbations of the same input. However, most existing approaches apply uniform consistency constraints across all pixels, regardless of the model's confidence in its predictions. This design choice may be inadequate, since ambiguous regions — such as lesion boundaries, low-contrast tissues or areas are inherently uncertain. Enforcing strong consistency constraints in these regions may propagate unreliable signals, mislead optimization, and amplify errors. To address this issue, inspired by [19], we propose an Uncertainty-aware Scale Consistency (USC) loss, which selectively applies stronger consistency supervision to confident regions while down-weighting ambiguous or noisy areas.

More specifically, given an input image $x$, we generate two perturbed versions $x^{(1)}$ and $x^{(2)}$ by applying random scale transformations. We use the same segmentation network $f_\theta(\cdot)$ to produce the corresponding probability maps:

$$p^{(1)} = f(x^{(1)}), p^{(2)} = f(x^{(2)}) \qquad (1)$$

To avoid enforcing consistency in ambiguous regions, we estimate the pixel-wise uncertainty using binary entropy:

$$H(p) = -[p\log(p+\epsilon) + (1-p)\log(1-p+\epsilon)] \qquad (2)$$

where $p \in [0,1]$ is the predicted probability after the sigmoid function, and $\epsilon$ is a small constant for numerical stability. The final uncertainty map is obtained by averaging the two entropies calculated from $p^{(1)}$ and $p^{(2)}$:

$$Uncertainty = \frac{1}{2}(H(p^{(1)}) + H(p^{(2)})) \qquad (3)$$

We then convert uncertainty map into a confidence weight map:

$$W = exp(-Uncertainty) \qquad (4)$$

With the confidence weight map $W$, the Uncertainty-aware Scale Consistency loss is formulated as:

$$\mathcal{L}_{usc} = \frac{1}{N}\sum_{i=1}^{N} W_i \cdot (p_i^{(1)} - p_i^{(2)})^2 \qquad (5)$$

Where $N$ is the number of pixels and $i$ indexes each pixel. This loss function enforces agreement between predictions under scale perturbations.

Such that pixels with lower uncertainty (i.e., higher confidence) receive larger weights. In this way, the consistency constraint is mainly enforced on regions where the model is already confident about its predictions, such as the interior of foreground or background areas. These regions are more reliable and less affected by annotation noise, so enforcing strong consistency here provides a stable training signal. Conversely, regions with high uncertainty—often located near object boundaries or in noisy areas— are naturally down-weighted. Forcing strict consistency in these ambiguous regions may misguide optimization, since the model has not yet reached a confident decision. By weighting the consistency loss according to entropy, the network can focus on consolidating its confident predictions while avoiding over-regularization in uncertain areas. The key idea of the proposed USC loss is that consistency regularization should emphasize stable knowledge rather than enforce agreement on unreliable regions. Instead of treating all pixels equally, USC adaptively weights predictions according to their uncertainty—encouraging consistency in confident areas while avoiding the propagation of noise from ambiguous lesion boundaries. This strategy leads to more robust and effective learning under weak supervision.

Nevertheless, if the segmentation results of $p^{(1)}$ and $p^{(2)}$ are transformed into box shape predictions, these predictions should be aligned with the ground-truth tight bounding box. To achieve this, the box alignment loss can be defined as follows:

$$\mathcal{L}_{boxalign} = BCEDiceLoss\left(Cat\left(p^{(1)}{}_{box}, p^{(2)}{}_{box}\right), Cat\left(GT_{box}, GT_{box}\right)\right) \tag{6}$$

where $p^{(1)}{}_{box}$ and $p^{(2)}{}_{box}$ denote the box-shaped predictions derived from $p^{(1)}$ and $p^{(2)}$, respectively. $GT_{box}$ represents the ground-truth tight bounding box.

### 3.4 Pseudo-label Refinement via UA-FGEPM

Although SAM2-generated pseudo labels are considerably more accurate than coarse labels derived solely from extreme points, they still exhibit substantial discrepancies when compared with the ground truths. If these pseudo labels are directly and persistently employed as the supervisory signal, the segmentation performance tends to saturate and cannot be further improved. This limitation highlights the necessity of an iterative training strategy, wherein pseudo labels are dynamically refined throughout the learning process.

To enhance the reliability and structural alignment of pseudo-label refinement in weakly supervised segmentation, we propose an Uncertainty-Aware Feature-Guided Extreme Point Masking (UA-FGEPM) algorithm by integrating Monte Carlo dropout-based uncertainty estimation [31] into the path cost construction. This module aims to regularize edge propagation using both gradient and uncertainty cues under a unified cost formulation.

More specifically, given an input image $x$, the segmentation model $f_\theta(\cdot)$ is evaluated $T$ times under stochastic dropout, yielding a set of feature predictions given by:

$$featureset = \{f_\theta^{(t)}(x)\}_{t=1}^{T} \tag{7}$$

The pixel-wise mean feature map $\mu(x)$ and variance map $\sigma^2(x)$ are computed as:

$$\mu(x) = \frac{1}{T}\sum_{t=1}^{T} f_\theta^{(t)}(x) \tag{8}$$

$$\sigma^2(x) = \frac{1}{T}\sum_{t=1}^{T} (f_\theta^{(t)}(x) - \mu(x))^2 \tag{9}$$

where $\mu(x)$ captures the stable structural representation while $\sigma^2(x)$ serves as a Bayesian approximation of epistemic uncertainty.

The mean feature map $\mu(x)$ is then processed with a Sobel operator to obtain the structural gradient map:

$$G(x) = \sqrt{G_x(\mu(x))^2 + G_y(\mu(x))^2} \tag{10}$$

where $G_x$ and $G_y$ are computed using horizontal and vertical Sobel kernels, respectively.

Simultaneously, the variance map $\sigma^2(x)$ will be normalized into the range [0,1], denoted as uncertainty map $U(x)$.

The final cost map is then defined as:

$$\text{Cost}(x) = \frac{1}{G(x) + \alpha \cdot U(x) + \varepsilon} \tag{11}$$

This expression can also be extended to compute the pixel-wise cost across the entire feature map.

Consequently, refined lesion contour is generated by connecting the four extreme points through minimum-cost paths on the constructed cost map. Specifically, we employ the route_through_array function from scikit-image [32], which implements an efficient variant of Dijkstra's algorithm over an 8-connected grid. The edge-wise traversal cost between adjacent pixels $v_i$ and $v_j$ can be defined as:

$$\text{Cost}(v_i \to v_j) = \frac{Cost(v_i) + Cost(v_j)}{2} \tag{12}$$

This design is motivated by two empirical facts: (1) true object boundaries typically exhibit strong gradients; (2) predictive uncertainty tends to peak near boundaries due to semantic ambiguity and prediction noise [33].

By minimizing the cost function, the derived paths are guided to follow regions that are consistent with structural edges while simultaneously incorporating model uncertainty. Based on these paths, the refined contour together with its enclosed area is designated as the foreground, whereas the remaining regions are assigned as the background, thereby yielding the refined pseudo label.

At intervals of K epochs, the pseudo labels are dynamically updated using the proposed UA-FGEPM algorithm, and the newly generated pseudo labels will replace those from the previous stage. The segmentation network is subsequently trained under the supervision of these progressively refined pseudo labels, using a hybrid loss function that combines Binary Cross Entropy and Dice loss. This iterative refinement mechanism allows the network to self-train through progressively improved supervision.As the model predictions become more accurate, the quality of pseudo labels also improves, thereby forming a mutually reinforcing loop between model optimization and label refinement. Upon completion of the iterative training process, the optimized network parameters are saved and employed for evaluation on the held-out testing sets. The loss function employing pseudo labels as the supervisory signal can be formulated as follows:

$$\mathcal{L}_{pl} = \frac{1}{2}\left(BCEDiceLoss(p^{(1)}, pseudolabel) + BCEDiceLoss(p^{(2)}, pseudolabel)\right) \tag{13}$$

Finally, our method is trained with a total loss combining the three losses mentioned above:

$$\mathcal{L}_{total} = \mathcal{L}_{boxalign} + \lambda_1 \cdot \mathcal{L}_{usc} + \lambda_2 \cdot \mathcal{L}_{pl} \tag{14}$$

Where $\lambda_1$ and $\lambda_2$ are hyperparameters to balance different losses.

## 4. Experiments and Results

### 4.1 Datasets

Two publicly available ultrasound datasets are employed to comprehensively evaluate the effectiveness of the proposed method, namely the Breast UltraSound Images (BUSI) dataset [34] and the Ultrasound Nerve Segmentation (UNS) dataset [35]. The BUSI dataset originally consists of normal, benign, and malignant images with corresponding expert-annotated masks. Following the setting in UNeXt [36], we consider only benign and malignant cases for the binary lesion segmentation task. The UNS dataset contains ultrasound images where the nerve regions must be delineated to facilitate accurate placement of a patient's pain management catheter. In line with [37], a total of 2,323 image–mask pairs are adopted for binary segmentation.

To ensure a fair and reliable evaluation, we adopt a five-fold cross-validation strategy on both datasets, since no official test set is provided. Specifically, the dataset is divided into five subsets of comparable size. A standard five-fold cross-validation protocol is then applied, where each subset is used as the validation set once, and the mean and standard deviation across the five folds are reported. This protocol eliminates the bias of a single split and provides a more robust estimate of the model's generalization ability.

It is worth noting that both datasets pose significant challenges for weakly supervised segmentation. The original ultrasound images are grayscale, and the lesion or nerve regions often exhibit intensity distributions highly similar to the surrounding tissues, leading to blurred and indistinct boundaries. In particular, the UNS dataset is notoriously difficult, as even expert physicians may struggle to accurately localize the nerves. These properties highlight the necessity of developing robust and uncertainty-aware weakly supervised segmentation methods.

### 4.2 Evaluation Metrics

To evaluate the segmentation performance of different methods, we adopt Intersection over Union (IoU) and Dice coefficient (equivalent to the F1 score) as quantitative metrics to measure the overlap between the predicted masks and the ground truths. For each dataset, the IoU and Dice are first computed on every image, and the results are averaged within each fold. Finally, we report the overall performance as the mean and standard deviation across the five folds.

4.3 Quantitative Comparison with Other Methods

Table 1 presents the quantitative comparison between our proposed approach and several representative weakly supervised segmentation methods. Fully supervision denotes training with pixel-level ground truths provided by the original dataset, which naturally achieves the best performance and serves as the upper bound for weakly supervised methods.

For comparison, we include five existing weakly supervised approaches: BoxInst [38], WeakPolyp [11] and BTBB [12], which rely on bounding-box supervision; Deep Geodesics [7] and PAplusNet [39], which are based on extreme-point annotations. Specifically, BoxInst minimizes the discrepancy between the bounding-box projections of the predicted masks and the ground truths. WeakPolyp learns segmentation purely from bounding-box annotations by introducing a mask-to-box transformation to mitigate label noise and a scale-consistency loss to provide dense supervision. BTBB exploits bounding boxes of varying precision levels as supervisory signals. Deep Geodesics were originally designed to connect six extreme points in 3D objects, and we adapt the implementation for 2D medical image segmentation. PAplusNet introduces a composite loss based on multi-level labels generated from four extreme points.

As shown in Table 1, our method consistently outperforms the bounding-box based approaches (BoxInst, WeakPolyp and BTBB) as well as the extreme-point based approaches (Deep Geodesics and PAplusNet). Notably, our method surpasses the fully supervised performance on the BUSI dataset, while on the UNS dataset it achieves results that are highly comparable to the fully supervised upper bound, which demonstrates both the effectiveness and practicality of our weakly supervised framework.

TABLE I. QUANTITATIVE COMPARISON RESULTS ON BUSI AND UNS DATASETS.

| Method | BUSI | | UNS | |
| --- | --- | --- | --- | --- |
| | IoU(%) | Dice(%) | IoU(%) | Dice(%) |

| Method | IoU | Dice | IoU | Dice |
|---|---|---|---|---|
| Fully supervision | 66.71 ± 3.15 | 76.01 ± 2.68 | 67.54 ± 1.11 | 79.08 ± 0.97 |
| BoxInst [38] | 59.53 ± 4.13 | 70.11 ± 3.33 | 63.09 ± 1.04 | 75.67 ± 0.95 |
| WeakPolyp [11] | 65.30 ± 3.14 | 74.96 ± 2.74 | 66.36 ± 1.01 | 78.30 ± 1.00 |
| BTBB [12] | 39.82 ± 2.06 | 51.53 ± 2.68 | 48.62 ± 1.63 | 63.59 ± 1.50 |
| Deep geodesics [7] | 57.91 ± 3.71 | 65.37 ± 2.89 | 62.33 ± 1.09 | 73.59 ± 0.96 |
| PAplusNet [39] | 53.83 ± 3.29 | 66.10 ± 2.92 | 52.63 ± 5.82 | 67.40 ± 4.96 |
| (Ours) | 67.70 ± 2.01 | 76.89 ± 1.85 | 67.39 ± 1.01 | 79.12 ± 0.96 |

4.4 Ablation Study

To further validate the contribution of each component in our framework, we conduct ablation studies, and the results are summarized in Table 2. The baseline model, trained with bounding boxes derived from extreme points, achieves relatively modest performance on both datasets. Incorporating SAM2 prompts leads to a substantial improvement (IoU: +7.43 pp on BUSI, +2.74 pp on UNS), confirming the effectiveness of foundation model knowledge for generating reliable initial pseudo labels. Adding the USC loss together with the box alignment loss yields further gains, with IoU reaching 66.05% on BUSI and 64.57% on UNS, which highlights the importance of explicitly regularizing scale-invariant and box-consistent predictions. When adopting the UA-FGEPM with iterative refinement, the IoU is further improved to 66.96% and 63.59% on BUSI and UNS, respectively, demonstrating the benefit of uncertainty-guided pseudo label refinement. Finally, combining all components achieves the best results, with our method reaching 67.70% IoU / 76.89% Dice on BUSI and 67.39% IoU / 79.12% Dice on UNS, surpassing all individual variants and approaching (or even exceeding) the fully supervised upper bound. These results clearly demonstrate the complementary advantages of each component and the overall effectiveness of our proposed framework.

TABLE II. ABLATION STUDIES ON BUSI AND UNS DATASETS.

| Strategy | BUSI | | UNS | |
|---|---|---|---|---|
| | IoU(%) | Dice(%) | IoU(%) | Dice(%) |
| Baseline | 56.99 ± 1.67 | 69.62 ± 1.28 | 57.03 ± 1.02 | 71.33 ± 0.93 |
| + SAM2 | 64.42 ± 2.17 | 74.57 ± 2.13 | 59.77 ± 0.65 | 73.47 ± 0.66 |
| SAM2+ USC Loss & Box Alignment Loss | 66.05 ± 1.95 | 75.60 ± 1.62 | 64.57 ± 1.08 | 77.10 ± 1.03 |
| SAM2+ UA-FGEPM & Iteration | 66.96 ± 2.33 | 76.37± 2.32 | 63.59 ±1.15 | 76.50 ± 0.96 |
| All | 67.70 ± 2.01 | 76.89 ± 1.85 | 67.39 ± 1.01 | 79.12 ± 0.96 |

## 4.5 Visualization of results

Fig.3 presents the qualitative comparison results obtained by different methods on representative samples from both datasets. As shown in the figure, our proposed method produces segmentation masks that are closest to the ground truths, even outperforming the fully supervised counterpart in terms of contour accuracy and region completeness. This superiority can be attributed to the iterative pseudo-label refinement guided by the four extreme points, which effectively corrects boundary inaccuracies and enforces spatial consistency during training. In contrast, the fully supervised model, though trained with dense annotations, lacks this iterative correction mechanism and occasionally yields slight contour drift, particularly along ambiguous lesion boundaries.

Moreover, other weakly supervised methods tend to either over-segment or include excessive false positives in background regions, particularly for the samples from the UNS dataset (i.e., the last three rows), where the nerve structures are elongated and low-contrast. Our framework, by combining SAM2-based initialization and MC-Dropout-driven uncertainty-aware refinement, maintains more stable and compact lesion segmentation results, suppressing noise and artifacts in ambiguous areas. Notably, in challenging cases with irregular or small lesions, our method achieves clearer boundaries and better shape consistency, indicating that the proposed uncertainty-guided FGEPM refinement contributes to both robustness and generalization.

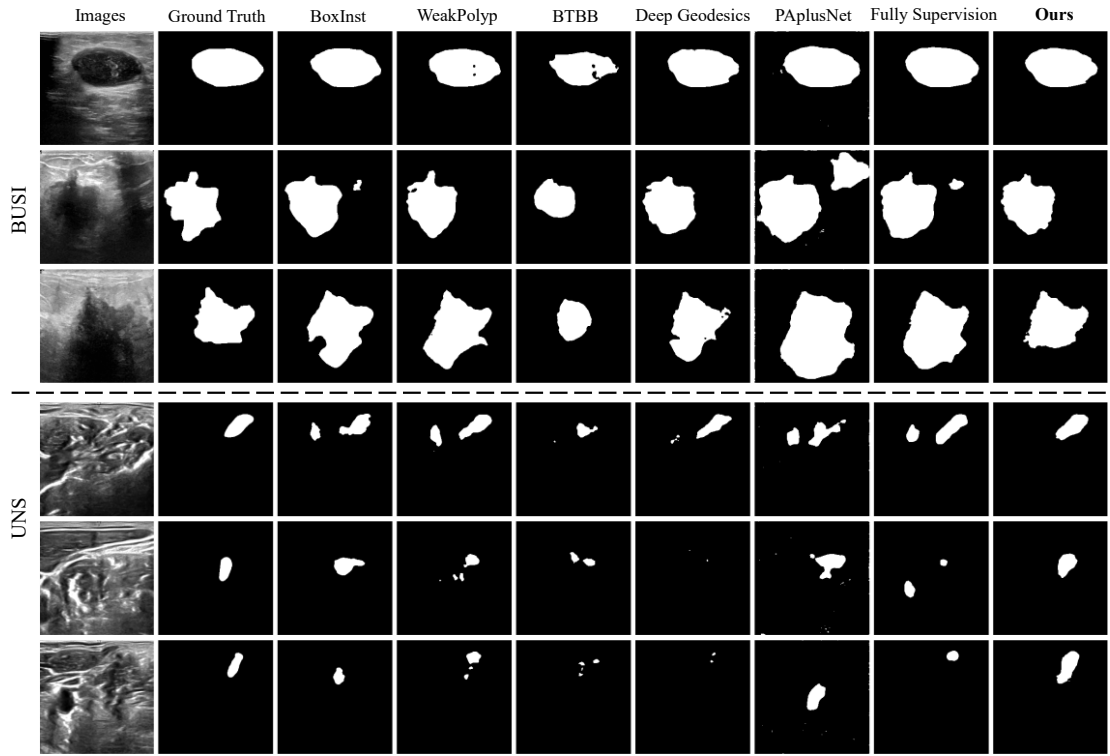

Fig.3. Visualization comparison results with different methods.

Fig. 4 presents the qualitative results of the ablation study. The baseline is trained under the supervision of tight bounding boxes derived from the four extreme points. As shown in the figure, the segmentation masks produced by the baseline are coarse and often fail to capture the complete lesion regions. By incorporating SAM2 for initial pseudo-label generation, the lesion boundaries become more consistent with the actual contours. When the USC loss and box alignment loss are further introduced, the predicted regions align more tightly with the bounding-box constraints, and background noise is significantly suppressed. Moreover, integrating the UA-FGEPM algorithm and iterative pseudo-label refinement further improves the segmentation accuracy, especially around ambiguous and low-contrast regions. Finally, when all modules are jointly applied, the segmentation results exhibit the most precise and complete boundaries, confirming the complementary benefits of each component and the overall effectiveness of our proposed framework.

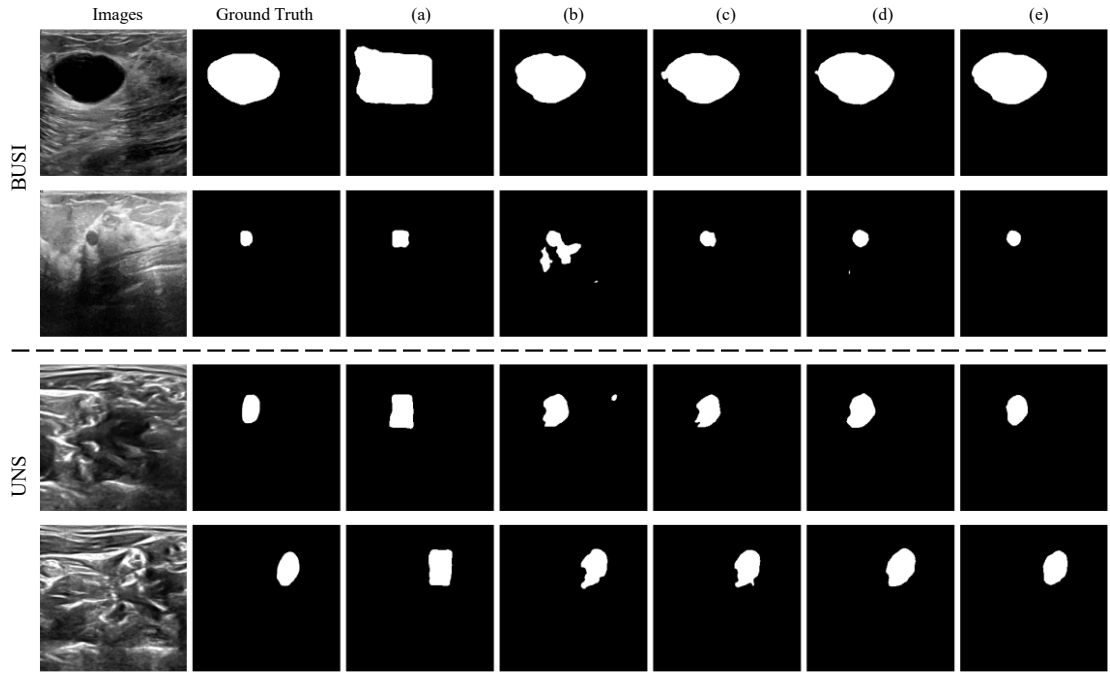

Fig.4. Visualization results of the ablation study. From left to right: original image, ground truth, (a) Baseline, (b) SAM2, (c) SAM2 + USC Loss & Box Alignment Loss, (d) SAM2 + MC-FGEPM & Iteration, and (e) All modules combined.

Fig. 5 presents the ablation studies on three key hyperparameters of our framework, including the number of Monte Carlo (MC) inference runs, the pseudo-label update frequency, and the dropout rate. As shown in Fig. 5(a), increasing the number of MC inference runs gradually improves segmentation performance, indicating more stable uncertainty estimation. However, the performance gain becomes marginal beyond 15 runs (IoU improves only 0.4 % from 15 to 20 runs, compared with 1% from 5 to 20 runs), while the computational cost of MC sampling increases linearly with the number of forward passes. Therefore, 20 runs are adopted as a practical balance between accuracy and efficiency. In Fig. 5(b), updating pseudo labels every 100 epochs yields the most stable performance. More frequent updates (e.g., 10 or 30 epochs) introduce unstable supervision due to immature pseudo labels, whereas overly delayed updates weaken the iterative refinement effect. Regarding the dropout rate in Fig. 5(c), a smaller probability ($p = 0.2$) achieves the highest mean IoU, suggesting that mild stochastic regularization enhances generalization without leading to underfitting.

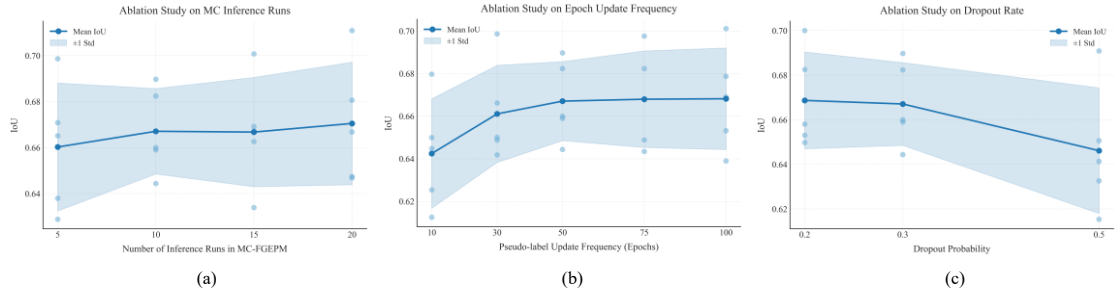

(a)  (b)  (c)

Fig.5. Ablation studies on key hyperparameters: (a) MC inference runs, (b) update frequency, (c) dropout rate.

To further illustrate the effectiveness of our iterative refinement strategy, Fig. 6 visualizes the evolution of pseudo labels across different training stages. In each column, the predicted contours (blue → cyan → green → red) are overlaid with the ground-truth boundary (white). The initial SAM2 pseudo labels can roughly localize the lesion area, but their boundaries are coarse and lack fine structural details. As the iterative training proceeds, the generated pseudo labels become progressively cleaner and more spatially consistent, exhibiting finer alignment with the true lesion boundary. The pseudo labels gradually converge to the ground-truth boundary, confirming the stability and effectiveness of the proposed refinement mechanism.

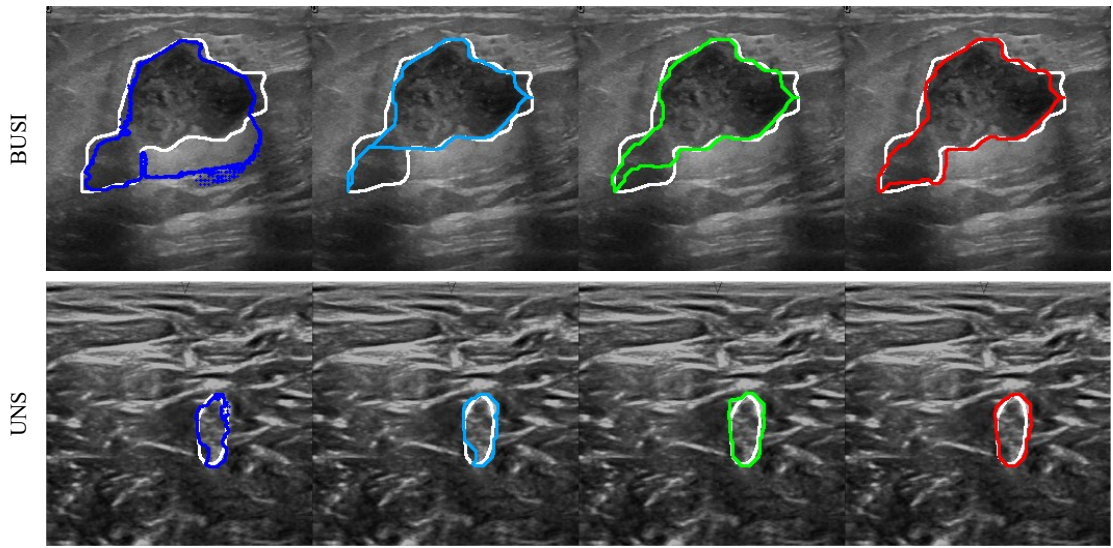

Fig.6. Visualization of pseudo-label evolution over training iterations. Each column represents the overlaid contours of the ground truth (white) and pseudo labels obtained at different stages: initial SAM2 (blue), epoch 100 (cyan), epoch 200 (green), and epoch 300 (red). The pseudo labels gradually converge to the ground-truth boundary.

5.Discussion and Conclusion

The proposed framework achieves strong and stable segmentation performance on both BUSI and UNS datasets, demonstrating that reliable supervision can be derived from only four extreme points. This confirms the effectiveness of combining SAM2-generated priors with the proposed uncertainty-aware pseudo-label refinement strategy. By progressively improving the quality of pseudo labels and integrating multiple weak cues, the framework maintains accurate boundary localization while greatly reducing annotation effort. Compared with previous weakly supervised methods that rely solely on bounding boxes or extreme points, our approach offers more consistent and anatomically meaningful predictions. These results suggest that the proposed design provides a practical and scalable solution for weakly supervised medical image segmentation.

In future work, we intend to extend our framework to other medical imaging datasets, such as dermoscopic and CT scans, to further evaluate its robustness and generalization in diverse clinical scenarios. Furthermore, we plan to explore more interactive segmentation strategies. For cases where the segmentation performance on certain test images is suboptimal, additional weak cues such as additional clicks, scribbles, or boundary hints can be interactively added to guide refinement. Recent studies[40][41] have shown that incorporating limited user interaction after model inference can effectively correct local boundary errors while maintaining low annotation cost.